# Selective Unsupervised Feature Learning with Convolutional Neural Network (S-CNN)

Amir Ghaderi
Department of Computer Science and Engineering
University of Texas at Arlington
Arlington, USA
amir.ghaderi@mavs.uta.edu

Vassilis Athitsos
Department of Computer Science and Engineering
University of Texas at Arlington
Arlington, USA
Athitsos@uta.edu

*Abstract*— Supervised learning of convolutional neural networks (CNNs) can require very large amounts of labeled data. Labeling thousands or millions of training examples can be extremely time consuming and costly. One direction towards addressing this problem is to create features from unlabeled data. In this paper we propose a new method for training a CNN, with no need for labeled instances. This method for unsupervised feature learning is then successfully applied to a challenging object recognition task. The proposed algorithm is relatively simple, but attains accuracy comparable to that of more sophisticated methods. The proposed method is significantly easier to train, compared to existing CNN methods, making fewer requirements on manually labeled training data. It is also shown to be resistant to overfitting. We provide results on some well-known datasets, namely STL-10, CIFAR-10, and CIFAR-100. The results show that our method provides competitive performance compared with existing alternative methods. Selective Convolutional Neural Network (S-CNN) is a simple and fast algorithm, it introduces a new way to do unsupervised feature learning, and it provides discriminative features which generalize well.

*Keywords; Deep Learning; Artificial Neural Networks; Classification and Clustring*

## I. Introduction

A popular method in machine learning is Convolutional Neural Networks (CNNs). CNN had was of high interest to the research community in the 1990s, but after that its popularity receded compared to the Support Vector Machines (SVMs) [1]. One of the reasons was the relatively lower computational demands of SVMs. Training CNNs requires significantly more computational power and time than training SVMs.

With increased availability of powerful GPU processing, and using several improvements in network structure, Krizhevsky et al. [2] used CNNs to achieve the highest image classification accuracy on ImageNet Large Scale Visual Recognition Challenge(ILSVRC) [3]. After that result, CNNs have become widely popular in the computer vision and pattern recognition community, and have been applied to a variety of classification problems, including detection and localization [4]. CNNs have achieved the best results for detection on the PASCAL VOC dataset [1], and for classification on the Caltech-256 [5] and Caltech-101 datasets [5] [6]. Based on such results, CNNs have emerged as a leading method for supervised learning.

At the same time, a weakness of supervised learning using CNNs is the need for much larger amounts of labeled training data, compared to alternative methods. Acquiring a large number of labeled instances requires oftentimes significant time spent by humans to provide the labels, and significant costs. Furthermore, when training instances are labeled by humans, errors and inconsistency in labeling become an issue, especially when labeling large scale datasets. On the other hand, in many settings it is easy to obtain vast amounts of unlabeled data, making unsupervised learning an attractive alternative, provided of course that unsupervised learning can attain satisfactory accuracy.

In this paper, we propose an algorithm that learns features using CNNs that train on unlabeled data. We evaluate this algorithm on the STL, CIFAR-10, CIFAR-100 datasets, obtaining competitive performance compared to other methods.

In Figure 1 we show the overview of algorithm. Selective search finds the important parts of the object. Then CNN learns the features to classify those important parts. At the final step, an SVM is trained on the features. The following sections describe each of these components in detail.

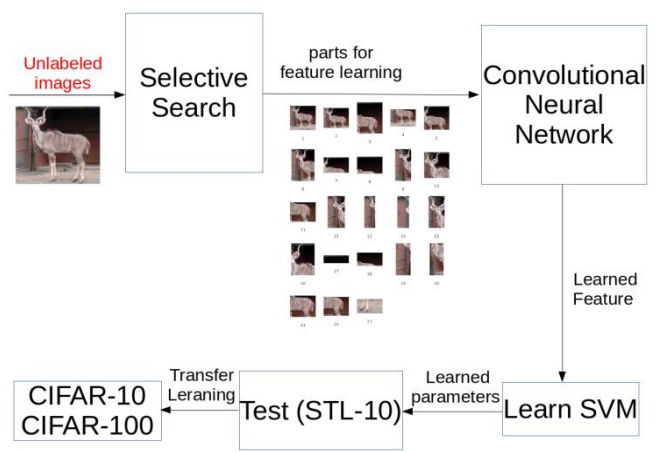

Figure 1: Overview of the algorithm

## II. RELATED WORK

CNN typically consist of different types of layers, with each layer performing some specialized functionality. Examples of such types of layers are convolutional layers, rectifier layers(max(0,x)) (also known as ReLU layers), max-pooling layers for reducing the number of inputs, and normalization layers [2]. The speed of training Deep CNN with ReLUs is much higher than the speed of training ReLUs with tanh units [2]. In fully connected layers, each element is calculated based on the values of all components of the input. The last layer calculates the loss function of the network. The main role of training is on the convolutional layers, and classification is performed by the fully connected layers. After training a CNN, instead of performing classification using the fully connected layers, one can feed features from the last convolution layer into an SVM classifier.

CNNs can be combined with both supervised and unsupervised methods in an end-to-end system. In supervised methods, data augmentation can be used to increase the number of instances for training, so as to reduce overfitting [2]. Coates et al. [7] point out that the effect of certain factors, such as the number of hidden nodes, may be more vital for performance than the depth of the model. In [8], researchers use the temporal slowness constraint with and employ a linear autoencoder in order to learn features from video. In the category of unsupervised methods, Bo et al. propose the hierarchical matching pursuit (HMP) method, which uses sparse coding and learns hierarchical feature representations in an unsupervised manner on depth data [9]. Unsupervised feature learning is used by Netzer et al. for recognizing digits cropped from street view images. Features invariant to transformations are learned by Sohn et al. [10]. Le et al. [11] have trained features robust to translation, scaling, and rotation for face detection using a deep sparse auto encoder on a large dataset, without having to label images.

## III. METHOD

Object detection in many methods is based on exhaustive search for specific object types. Alternatively, some methods output possible locations of objects, without being trained to detect specific types of objects. Such methods include objectness [12], selective search [13], and category-independent object proposals [14]. Selective search identifies potential object locations which can be used for object recognition. It combines advantages of both exhaustive search and segmentation [15] and achieves relatively high speed compared to alternative methods. It uses the structure of the image for sampling, and it creates scores by merging low-level superpixels [16]. The goal of selective search is to find all locations in the image that have high probability to be an object. The output of selective search given an image is a set of bounding boxes, representing possible locations of objects.

As we mentioned earlier, annotating large sets of images can be an important bottleneck for training supervised methods, but large amounts of unlabeled data may be easy to obtain. E.g. in the STL dataset there are 100K unlabeled images. Let $x_i$ be an unlabeled image, that we give as input to the selective search algorithm. Selective search outputs a set $w_i$ of bounding boxes for $x_i$. We treat each bounding box as a subimage of $x_i$. Thus, set $w_i$ consists of many images $a_{ij}$, which are all subimages of $x_i$.

$$w_i = \{a_{ij} | a_{ij} \text{ is output of selective search with input } x_i\}$$

If selective search creates $T_i$ subimages from $x_i$ then $j = \{1,2,3 \ldots T_i\}$ and $w_i = \{a_{i1}, a_{i2}, \ldots a_{iT_i}\}$. Then, we assign training label i to all these images in set $w_i$. In other words, $w_i$ generates $T_i$ image-label pairs [ $a_{ij}$, i] for our training set. Intuitively, all subimages from the same original image $x_i$ are assigned i as their training label. Thus, training labels are assigned fully automatically, with no need for manual intervention.

Set T contains as elements the numbers $T_i$ of subimages extracted from all unlabeled images $x_i$. We have 100,000 unlabeled images in the STL dataset, so T has 100,000 members.

$$T = \{T_1, T_2, \ldots, T_{100000}\}$$

Suppose that we want to train a CNN to recognize C classes, where C is a user-specified parameter. We want to find the C members of T that contain the most elements. For reaching this goal we sort set T in descending order, and we put the indices in set TS.

$$TS = \text{indices of sorted T in descending order} = \{ts_1, ts_2, \ldots ts_{100000}\}$$

Note that TS stores indices of elements in T, not the elements themselves. So $T_{ts_1}$ is the maximum element of the T. We choose the top C indices of TS to train the CNN. In our experiments, we try C= 5000, 10000, 15000, 20000, 25000, 30000. Our goal is to train a CNN to discriminate between C classes, and to choose features that can discriminate among various types of objects. Therefore, the input for training the CNN is a set of images X and labels as below:

$$X = \{w_{ts_1}, w_{ts_2}, \ldots, w_{ts_C}\}$$

$$\text{labels for images in } w_{ts_i} = ts_i$$

The loss function which should be minimized is:

$$L(X) = \sum_{w_i \in X} \sum_{a_{ij} \in w_i} l(i, a_{ij}) \qquad [1]$$

$l(i, a_{ij})$ is the softmax loss based on the image $a_{ij}$ and the label i. In the following section we provide more details about the trained CNN.

## IV. EXPERIMENTS

For comparison to other methods, we evaluate performance on the STL-10 dataset [7], which has 10 classes, and the CIFAR-10, and CIFAR-100 datasets [17] that have 10 and 100 classes respectively. STL-10 contains 100,000 unlabeled data we use it as source of data for unsupervised feature learning.

We extract the surrogate classes for training the CNN from the unlabeled set of STL-10. Each image in the unlabeled STL set is given an input for selective search. The output images of the selective search have different sizes, which would cause features created in fully connected layers to have different numbers of elements. To deal with this problem there are two options. The first is resizing the images to P*P fixed size, where P is a preselected parameter. The second is to use images with different sizes at beginning of the network, and to use spatial pyramid pooling [18] at the last layer before the fully connected layers, so as to create fixed number of features in fully connected layers. Here we select the first option and resize the input images to 32*32.

We try two network architectures. The first one has three convolutional layers, each of them with 64, 128, and 256 filters respectively. The kernel size for the first convolutional layer is 5*5. We use stride 1 and padding 2 for this layer. An ReLU filter is after each convolutional layer. After the first and the second ReLU layer we have the max pooling layer. Here we have kernel size 3*3, stride 2, and zero padding. The third ReLU layer is followed by two fully connected layers with 512 and C neurons respectively, where C is the number of the class labels that are assigned automatically (see Section III). Note that C varies in different experiments, as described later. Dropout [19] is employed at the fully connected layers to reduce overfitting. At the end there is a softmax layer for calculating the loss function. We named this network 64-128-256_512.

The second network, which is larger than the first one, has three convolutional layers with 92, 256, and 512 filters, followed by a fully connected layer with 1024 neurons. We named this network 92-256-512_1024. The kernel size for the first convolutional layer is 5*5. We use stride 1 and padding 2 for this layer. Again, a Rectified Linear Unit (ReLU) is used after each convolutional layer. After the first ReLU layer there is a max pooling layer with kernel size 3*3, stride 2, and zero padding. The second convolutional layer is like the first one, except that it consists of 256 kernels instead of 92. The ReLU and pooling layers applied to second convolutional layer are the same as for the first layer. The third convolutional layer has 512 kernels. At the end we have two fully connected layers with 1024 and C neurons, where again C is the number of classes and is different in each experiment. As in the first network, we have a softmax layer at the end for calculating the loss function. Figure 2 shows the second network in details. The figure is created by NVIDIA Deep Learning GPU Training System (DIGITS). We implement CNNs based on the caffe framework [20].

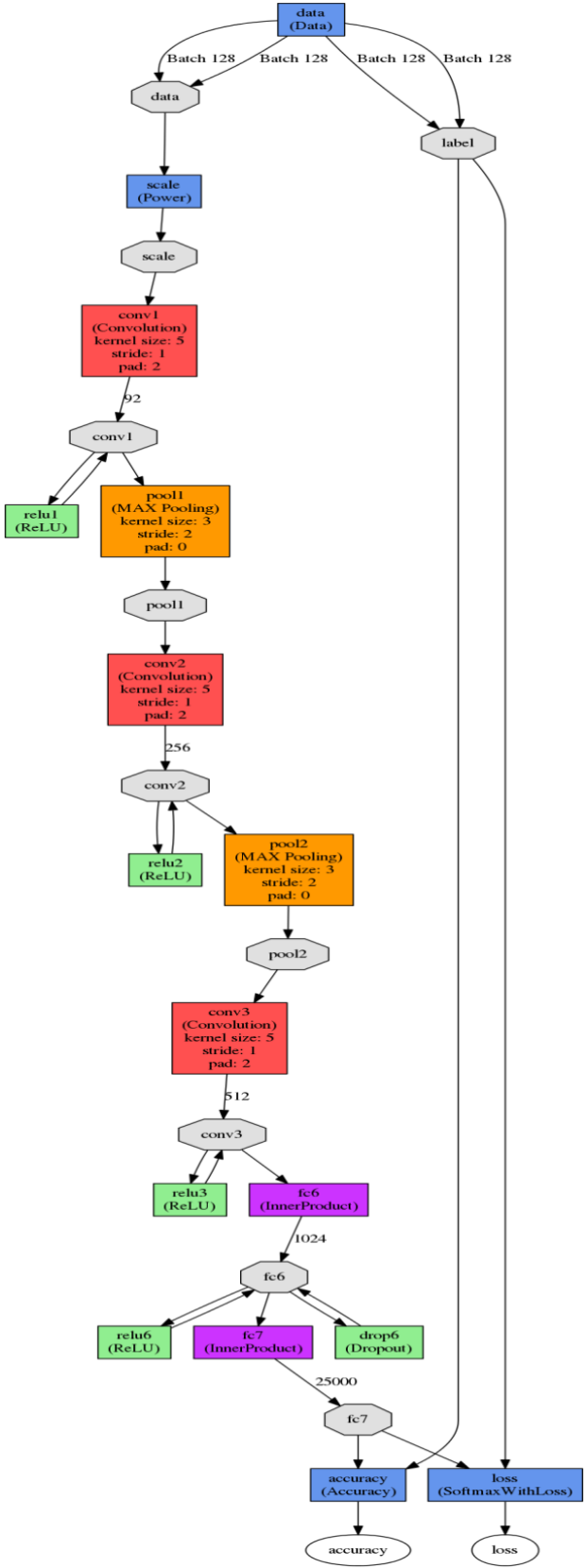

Figure 2: Architecture of network 92-256-512_1024

For each dataset, each image of the test set of that dataset is given as input to the network. Then, we compute the output of all the network layers expect the top softmax one. We use the pooling method which is usually used for the STL-10 dataset. 4-quadrant max-pooling, to obtain 4 values per feature map. This is the standard procedure for STL-10 [21]. We use the pooled features for training a one-vs-all linear support vector machine (SVM). To train the SVM we use the standard training and testing protocols for each dataset. For the STL dataset, we use the 10 predefined folds for training the SVM, and final accuracy is calculated as the average accuracy over the 10 splits. Code is available at http://vlm1.uta.edu/~amir/s-cnn.

Here we investigate the impact of different parameters on the results. We run different experiments by varying the number of classes, the network structure, and the dataset.

*A. Number of classes*

As described in Section III, parameter C is the number of classes that are assigned in an automatic manner, so as to train the CNN. We experimented with C equal to 5K, 10k, 15K, 20K, 25K, and 30K. A larger C can increase accuracy, because the neural network receives more training data. At the same time, when C is too large, the network can be fed with conflicting data (since class labels are assigned automatically) and not converge.

Table 1 shows the accuracy obtained on the STL dataset for different values of C. Indeed we notice that accuracy improves as C increases from 5000 to 20000, and then it starts decreasing. The best values of C are in the range between 20K and 25K.

Table 1: Accuracy percentages on the STL dataset using different values of C (number of classes).

| #classes | CNN | SVM(linear) |
|---|---|---|
| 5000 | 64-128-256_512 | 58.01 |
| 10000 | 64-128-256_512 | 58.10 |
| 15000 | 64-128-256_512 | 58.29 |
| 20000 | 64-128-256_512 | 61.04 |
| 25000 | 64-128-256_512 | 60.38 |
| 30000 | 64-128-256_512 | 58.87 |

*B. Generality of features*

We have also used the features learned on the STL dataset for recognition on the CIFAR-10 and CIFAR-100 datasets. These two are popular datasets, used by several researchers. Both datasets are split into a training set and a test set. In contrast to the STL dataset, the CIFAR datasets do not have any unlabeled data. We do not use their training set to learn features by CNN, using instead the trained features from the STL dataset. The results are comparable to other methods which use the CIFAR training sets directly. Table 2 shows the results for classification on CIFAR-10 and CIFAR-100 with learned features from STL-10.

Table2: Classification accuracy percentages on the CIFAR-10 and CIFAR-100 datasets

| Method | CIFAR-10 | CIFAR-100 |
|---|---|---|
| S-CNN(64-128-256_512) | 72.68 | 47.70 |
| S-CNN(92-256-512_1024) | 75.17 | 51.27 |
| [7] | 79.7 | 70.2 |
| [22] | - | 54.32 |

*C. Different network architectures*

We have conducted additional experiments to investigate the impact of the network architecture on classification performance. Since we established the best range for parameter C (number of classes) is 20K-25K, we decided to run two different architectures for the neural network, trained with C equal to 20000 and 25000. The details of the architectures are explained at the beginning of Section IV. The 92-128-512_1024 network has more parameters to learn and more power to discriminate between classes relative to the 64-128-256_512 network. We only change the parameters of the layers, and the number of layers is fixed for both network architectures. The 64-128-512_1024 network with 25K classes has 61.94 percent accuracy on STL test set. It shows that this architecture has more power for creating more distinguishing features. Classification accuracy improves with increasing network size. This is evidence that our algorithm works well with larger networks and avoids overfitting. The results of these experiments with different neural network architecture on the STL-10 dataset are shown in table 3.

Table 3: Accuracy percentages of different architectures on the STL dataset.

| Architecture | #classes | Accuracy |
|---|---|---|
| 64-128-256_512 | 20000 | 61.04 |
| 64-128-256_512 | 25000 | 60.38 |
| 92-256-512_1024 | 20000 | 60.36 |
| 92-256-512_1024 | 25000 | 61.94 |

*D. Comparison to other methods*

In Table 4 we compare the results of our algorithm with other learning methods on the STL-10 dataset. Our approach appears to be competitive with the others, despite the fact that our model only uses 3 convolutional layers and requires learning only few parameters. Note that better result than ours which reported in the table have been obtained by using external data, achieving an accuracy rate of 70.10% on STL-10 [23]. In that work, knowledge gained from previous optimizations is transferred to new tasks in order to find optimal hyperparameter settings more efficiently. We find it particularly promising that our results are more accurate than those of [7], [8], [10], and [24].

Table 4: Classification accuracy percentages on the STL-10 dataset

| Our method | [23] | [8] | [24] | [10] | [7] |
|---|---|---|---|---|---|
| 61.94 | 70.10 | 61.0 | 60.1 | 58.70 | 51.5 |

## V. DISCUSSION AND FUTURE WORK

In this paper we have proposed a new method for unsupervised feature learning, tailored for image classification in large datasets. We show that results are compatible to previously proposed methods, while our results use a simpler architecture and no data augmentation or use of external data. Also, the features learned on the STL-10 dataset are tested on the CIFAR-10 and CIFAR-100 datasets, and results show that the learned features generalize well and can extend to other sets of data.

There are several interesting directions for improvements. One such direction is trying bigger and deeper architectures for CNN. Using CNNs with more layers may learn more powerful features for distinguishing among different objects. Another interesting direction is to try learning features from a bigger dataset, with more images and classes than the STL-10 unlabeled dataset, to see if that would lead to learning better features.